\pdfoutput=1
\documentclass[11pt]{article}

\usepackage[final]{acl}

\usepackage{times}
\usepackage{latexsym}

\usepackage[T1]{fontenc}
\usepackage[utf8]{inputenc}

\usepackage{microtype}

\usepackage{inconsolata}

\usepackage{todonotes}
\usepackage{amsmath}
\usepackage{cleveref}
\crefname{section}{\S}{\S}
\crefname{table}{Table}{Tables}
\crefname{figure}{Fig.}{Figs.}
\crefname{algorithm}{Alg.}{}
\crefname{ALC@unique}{Line}{Lines}
\crefname{equation}{Eq.}{Eqs.}
\crefname{appendix}{App.}{Apps.}
\crefformat{section}{\S#2#1#3}

\usepackage[normalem]{ulem}
\usepackage{graphicx}
\usepackage{multirow}
\usepackage{hhline}
\usepackage{arydshln}

\title{UMBCLU at SemEval-2024 Task 1: Semantic Textual Relatedness with and without machine translation}

\renewcommand*{\thefootnote}{\fnsymbol{footnote}}

\author{Shubhashis Roy Dipta \footnotemark[2] \and Sai Vallurupalli \footnotemark[2] \\
\texttt{\{sroydip1,kolli\}@umbc.edu} \\
    Department of Computer Science and Electrical Engineering \\
    University of Maryland, Baltimore County \\
    Baltimore, MD 21250 USA \\
  }

\begin{document}

\maketitle
\def\thefootnote{†}\footnotetext{These authors contributed equally to this work}

\renewcommand{\thefootnote}{\arabic{footnote}}

\begin{abstract}
The aim of SemEval-2024 Task 1, ``Semantic Textual Relatedness for African and Asian Languages'' is to develop models for identifying semantic textual relatedness (STR) between two sentences using multiple languages (14 African and Asian languages) and settings (supervised, unsupervised, and cross-lingual). %Leveraging several types of LLMs, we developed models for identifying STR in the supervised (Subtask A) and the cross-lingual (Subtask C) settings. ).
Large language models (LLMs) have shown impressive performance on several natural language understanding tasks such as multilingual machine translation (MMT), semantic similarity (STS), and encoding sentence embeddings. Using a combination of LLMs that perform well on these tasks, we developed two STR models, \textit{TranSem}  and  \textit{FineSem}, for the supervised and cross-lingual settings.  We explore the effectiveness of several training methods and the usefulness of machine translation.  We find that direct fine-tuning on the task is comparable to using sentence embeddings and translating to English leads to better performance for some languages.
In the supervised setting, our model performance is better than the official baseline for 3 languages with the remaining 4 performing on par. In the cross-lingual setting, our model performance is better than the baseline for 3 languages (leading to $1^{st}$ place for Africaans and $2^{nd}$ place for Indonesian), is on par for 2 languages and performs poorly on the remaining 7 languages. 
\end{abstract}

\section{Introduction}

The objective of the SemEval 2024 Task 1 is to build and evaluate models capable of identifying relatedness between a sentence pair.  Sentence pairs from 14 African and Asian languages belonging to 5 language groups are annotated for relatedness and released for model development.  The task is divided into 3 tracks targeting different types of model training: supervised (Track A), unsupervised (Track B), and cross-lingual (Track C). Each track targets a different subset of languages. Extensive details about the languages, language groups, and the data collection process are provided in the task description paper \cite{ousidhoum2024semrel2024}.  A detailed description of the shared task, tracks, and datasets are provided in the shared task description paper~\cite{ousidhoum-etal-2024-semeval}.

Semantic relatedness helps with understanding language meaning \citep{Jarmasz2012RogetsTA,miller,elife,osgood} and is useful in many areas of natural language processing such as word-sense disambiguation \citep{wordsense}, machine translation \citep{BRACKEN_DEGANI_EDDINGTON_TOKOWICZ_2017} and sentence representation \citep{Reimers_2019, wang-etal-2022-just} which have numerous applications. Until recently, semantic relatedness has been mostly restricted to finding word relatedness \citep{Feng_Bagheri_Ensan_Jovanovic_2017,budanitsky}, leading to a lack of sentence-relatedness datasets. At the sentence level, relatedness has been limited to similarity, providing a restricted view of STR \citep{abdalla}. The current shared task aims to broaden the scope of sentence relatedness and extend it to several languages with the goal of encouraging model and resource development in these languages \citep{ousidhoum-etal-2024-semeval}.

%LLMs have \citet{Reimers_2019}
%In this paper, we describe our two models for supervised and cross-lingual tracks. 
Recent advancements in multi-lingual translation and the availability of models for obtaining high-quality sentence embeddings allowed us to explore  the effectiveness of machine translated data.  Using various sentence embedding models to encode data translated into English, we trained a model, \textit{TranSem}, to find the relatedness score between sentence pairs. Although the task requires the sentence pair to be from the same language, our model can handle sentences from two different languages.  Our second model, \textit{FineSem} directly fine-tunes a T5 model (T5 is already fine-tuned on the STS benchmark) on the STR task using both untranslated and translated data to explore the usefulness of translation. We use both these models to evaluate languages in Track A.  For Track C languages, we use a T5 model fine-tuned only on the english STR data. For evaluating  the English dataset in the cross-lingual track, we use a T5 model fine-tuned on the Spanish dataset. 

Our contributions to the STR task are as follows: We 1) develop unified models for STR to work with all languages.
    2)  participate in supervised and cross-lingual tracks.
    3) explore the usefulness of machine translation.
    4) explore data augmentation using machine translation.
 Our code is publicly available at \url{https://github.com/dipta007/SemEval24-Task8}

\section{System Overview}
After exploring models and datasets available in the languages we understand\footnote{Of the 14 languages, the authors are proficient in English, Hindi \& Telugu languages.}, we realized the dearth of resources available in these languages.  To leverage resources available in English, we translated the 13 non-English languages into English.   
%The result of our effort lead to two different systems for the supervised (track A) and cross-lingual (track C) training tracks. For track A, our goal was to develop a single model for all languages and explore the effectiveness of machine translation. For track C, our goal was to explore the effectiveness of transfer learning from English STS \& STR to STR in other languages.
%\subsection{Model for Track A}
%In this section, we describe our translation-based sentence embedding system, \textit{TransSem} for use in the supervised setting. Our goal was to translate sentences from the other languages to English and  train a single model on the translated data. 
Assuming the translated data accurately reflects the semantic meaning of the source language, the derived relatedness value from our model for a translated sentence pair should reflect the STR between the sentence pair in the source language. This section describes our machine translation process and models, \textit{TranSem} and \textit{FineSem}.  Besides using different training strategies, these models can use both translated and untranslated data.  
%We evaluate machine translation performance by comparing the performance of translated datasets with the English dataset, which does not undergo translation. %Our system for Subtask A is divided into two parts. The first is the translation (described on \cref{sec:translation}), and the second part is the main model (described on \cref{sec:model}). 

\subsection{Translation to English \& Data Augmentation} \label{sec:translation}
We use Meta's ``No Language Left Behind (NLLB) open-source models'' that provide direct high-quality translations for 200 languages with many low-resource languages~\citep{costa2022no}. We use  four of the translation models\footnote{
    \href{https://huggingface.co/facebook/nllb-200-3.3B}{facebook/nllb-200-3.3B},
    \href{https://huggingface.co/facebook/nllb-200-1.3B}{facebook/nllb-200-1.3B}, 
    \href{https://huggingface.co/facebook/nllb-200-distilled-1.3B}{facebook/nllb-200-distilled-1.3B}, 
    \href{https://huggingface.co/facebook/nllb-200-distilled-600M}{facebook/nllb-200-distilled-600M}
}
ranging from 600 million to 3.3 billion parameters.   Using each of the 4 models, we translated the training data for all languages in track A, except Amharic and Algerian Arabic, and obtained 4 translated datasets for each language.  None of the 4 models we used supported Amharic, Algerian Arabic or Punjabi. We decided against translating track C languages with 3 of 12 unsupported languages.  Using 4 different model translations gave us a 4-fold augmentation of the training data.  We translated the test data using only the largest model (\emph{facebook/all-200-3.3 B}) to obtain the best translation features. 

\subsection{\textit{TranSem} Model} \label{sec:model}
\begin{figure}[!t]
    \centering
    \includegraphics[width=\linewidth]{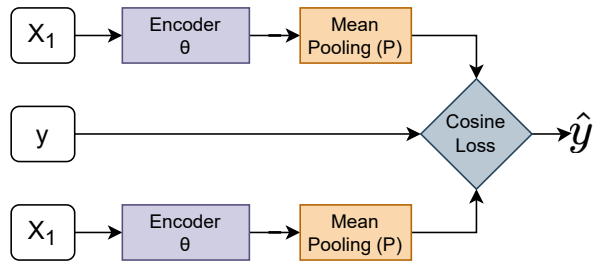}
    \caption{Overview of TranSem model architecture (Inspired by \citet{Reimers_2019}). The encoder ($\theta$) is shared, and the diamond box represents the loss function. The encoded sentence pairs ($x_1, x_2$) and the label ($y$) are the input to the cosine similarity loss.}
    \label{fig:main_model_A}
\end{figure}

Inspired by \citet{Reimers_2019}, we used the Siamese model architecture (shown in \cref{fig:main_model_A}). For a given pair of sentences ($x_1, x_2$) and their semantic relatedness score ($y$), we encode each sentence with a sentence encoder ($\theta$). The embeddings for the sentences go through a pooling operation ($P$) to produce sentence embeddings ($s_1, s_2$). The cosine similarity of the encoded embeddings is trained to match the semantic relatedness score using the mean-squared error loss:
\begin{equation}
    \mathcal{L} = MSE(\text{cos-sim}(P\theta (x_1), P\theta (x_2)), y)
\end{equation}

We experimented with several sentence encoding models for encoding our translated and augmented training dataset. We chose DistilRoberta\footnote{\href{https://huggingface.co/sentence-transformers/all-distilroberta-v1}{sentence-transformers/all-distilroberta-v1}} to submit results for the competition leaderboard based on our primary validation (details on \cref{sec:a1}, \cref{sec:a2}, and \cref{sec:a3}). This is a distilled version of RoBERTa \citep{liu2019roberta} fine-tuned on sentence-level datasets and suitable for clustering and semantic searches, which we further fine-tuned on our translated and augmented dataset. The sentence-t5-xl embedding model was chosen to compare the effectiveness of sentence embeddings as opposed to the direct fine-tuning used in the \textit{FineSem} model.  After experimenting with different pooling mechanisms of mean, max, and CLS tokens, we found that mean pooling works well for our setting. This aligns with earlier findings, which show that mean pooling produced encodings lead to better performance on downstream tasks. 

\subsection{\textit{FineSem} Model}

T5 \cite{t5_raffel} is a transformer model that uses transfer learning; the model trained on ``Colossal Clean Crawled Corpus'' is fine-tuned on a mixture of 8 downstream unsupervised and supervised tasks converting them into a unified text-to-text task setting. The T5 model is available in several sizes, of which we use the base, large, and XL models ranging from (660 million to 3 billion parameters).  One of the supervised tasks used in the T5 model training is the semantic textual similarity benchmark (STS-B) dataset trained as a regression classification problem.  We use the STS task setting to train on the track A STR training datasets using 3 different options:  separate  T5 models trained on individual languages, a single model trained on all 14 languages (without translation), and a single model trained on the translated and augmented dataset (for 12 languages).  These settings allow us to contrast the effectiveness of direct fine-tuning with the sentence embedding-based \textit{transem} model and the usefulness of machine translation.

From the T5 models fine-tuned on the individual languages, we use the English and Spanish fine-tuned models (we refer to these models as the English and Spanish models) for evaluating the cross-lingual Track C languages. We use the English model to evaluate development and test data from all languages except English and the Spanish model to evaluate the English data.   

\section{Experimental Setup}

\subsection{HyperParameters}
We train our models using AdamW \citep{loshchilov2017decoupled} optimizer with a learning rate of 1e-5 and weight decay of 0.01. We use an effective batch size of 32 (batch size 16 with 2 steps of gradient accumulation) (for TransSem) and a batch size of 16 (for FineSem). 

We train our \textit{transem} model infinitely with an early stopping patience of 10 on the validation Spearman Correlation score.  We train the \textit{finesem} model for 10 epochs (2 epochs for the model trained on the translated and augmented model)  and checkpoint the models at the end of every epoch.  We evaluate the dev sets for each language against these 10 checkpoints.  We evaluate the corresponding test data using the checkpoint which provides the best performance on the dev data for a language.  

\subsection{Infrastructure}
All experiments were conducted on an NVIDIA Quadro RTX 8000 with 48GB of VRAM and A100 80GB. We utilize the PyTorch Lightning library \footnote{\url{https://lightning.ai/}} for conducting the experiments and Weight \& Biases \footnote{\url{https://wandb.ai/}} for logging purposes (for the \textit{TransSem} Model) and HuggingFace Transformers (for the \textit{FineSem} Model). 

\section{Results}
We first report our results and analysis on Track A languages (\cref{res:a}), and then on Track C (\cref{res:c}).  We use the official baseline (\citet{ousidhoum-etal-2024-semeval}) that used LaBSE \citep{feng2020language} fine-tuned on the provided training dataset and refer to this model as $baseline$.  

\subsection{Track A Languages} \label{res:a}
 This section discusses our findings using various model settings with the \textit{TranSem} model.

\subsubsection{Effect of Batch Size} \label{sec:a1}
Comparing performance with various batch sizes (results are shown in \cref{tab:result_a_batchsize}), we show that our batch size selections are fairly good (32 for task \textit{TranSem} and 8 for \textit{FineSem}).

\begin{table}[!ht]
\centering
\resizebox{\linewidth}{!}{%
    \begin{tabular}{l|ccccccccccc}
        \multicolumn{1}{l}{} & \begin{tabular}[c]{@{}c@{}}Batch\\ Size\end{tabular} & \begin{tabular}[c]{@{}c@{}}A3\\eng\end{tabular} & \begin{tabular}[c]{@{}c@{}}A4\\hau\end{tabular} & \begin{tabular}[c]{@{}c@{}}A5\\kin\end{tabular} & \begin{tabular}[c]{@{}c@{}}A6\\mar\end{tabular} & \begin{tabular}[c]{@{}c@{}}A7\\ary\end{tabular} & \begin{tabular}[c]{@{}c@{}}A8\\esp\end{tabular} & \begin{tabular}[c]{@{}c@{}}A9\\tel\end{tabular} & avg \\ 
        \hhline{~===========}
        TranSem+ &   2 & .8102 & \textbf{.6857} & .6886 & \textbf{.8515} & .7376 & \textbf{.6519} & \textbf{.8342} & \textbf{.7514} \\
        TranSem+ &   4 & .5415 & .1355 & .0643 & .5394 & .3038 & .5998 & .4231 & .3725 \\
        TranSem+ &   8 & .8132 & .6519 & .6642 & .8348 & .7340 & .6355 & .8187 & .7360 \\
        TranSem+ &   16 & .8093 & .6377 & \textbf{.6997} & .8429 & \textbf{.7500} & .6462 & .8257 & .7445 \\
        TranSem+ &   64 & .8092 & .6589 & .6456 & .8271 & .7247 & .6234 & .8131 & .7289 \\
        TranSem+ &   128 & .8129 & .6787 & .6659 & .8417 & .7411 & .6396 & .8324 & .7446 \\
        TranSem+ &   256 & \textbf{.8152} & .6716 & .6589 & .8357 & .7197 & .6353 & .8189 & .7365
    \end{tabular}
}
\caption{The effect of batch size on $TranSem$ for different batch sizes $\{2, 4, 8, 16, 64, 128, 256\}$}
\label{tab:result_a_batchsize}
\vspace{-2mm}
\end{table}

\subsubsection{Effect of Encoder Pooling} \label{sec:a2}
 In \cref{table:result_a_poolling}, we compare performance using 3 different pooling operations.  We used mean pooling for the official results we submitted, as it showed good performance in most languages.
 
\begin{table}[!ht]
\centering
\resizebox{\linewidth}{!}{%
\begin{tabular}{l|ccccccccccc}
\multicolumn{1}{l}{} & \begin{tabular}[c]{@{}c@{}}Pool\\ ing\end{tabular} & \begin{tabular}[c]{@{}c@{}}A3\\eng\end{tabular} & \begin{tabular}[c]{@{}c@{}}A4\\hau\end{tabular} & \begin{tabular}[c]{@{}c@{}}A5\\kin\end{tabular} & \begin{tabular}[c]{@{}c@{}}A6\\mar\end{tabular} & \begin{tabular}[c]{@{}c@{}}A7\\ary\end{tabular} & \begin{tabular}[c]{@{}c@{}}A8\\esp\end{tabular} & \begin{tabular}[c]{@{}c@{}}A9\\tel\end{tabular} & avg \\ 
\hhline{~===========}
TranSem+ & CLS & \textbf{.8133} & \textbf{.6737} & .6655 & .8339 & .7376 & .6363 & \textbf{.8309} & .7416 \\
TranSem & Mean &  .8125 & .6403 & \textbf{.6807} & \textbf{.8406} & \textbf{.7448} & \textbf{.7211} & .8255 & \textbf{.7522} \\
TranSem+ & Max &  .7960 & .6157 & .5809 & .8227 & .6643 & .6075 & .7997 & .6981 \\
\end{tabular}
}
\caption{The effect of pooling on $TranSem$ using different pooling mechanisms (CLS Token, Mean, Max)}
\label{table:result_a_poolling}
\end{table}

\subsubsection{Effect of Sentence Embedding Models} \label{sec:a3}
In \cref{table:result_a}, we provide contrastive results with several sentence embedding models used in \textit{TranSem}. 
For official results, we submitted results from the distilroberta-v1 sentence embedding model (results for some of the languages are from the \textit{FineSem} model fine-tuned on individual STR training datasets).

\begin{table*}[!ht]
\centering
\resizebox{.98\textwidth}{!}{
\resizebox{\linewidth}{!}{%
    \begin{tabular}{|l|c|c|c|c|c|c|c|c|c|c|}
    \hline
    \begin{tabular}[l]{@{}c@{}}\\Model\end{tabular}&
    %\begin{tabular}[c]{@{}c@{}}A1\\arq\end{tabular}&\begin{tabular}[c]{@{}c@{}}A2\\amh\end{tabular}&
    \begin{tabular}[c]{@{}c@{}}A3\\eng\end{tabular} & \begin{tabular}[c]{@{}c@{}}A4\\hau\end{tabular} & \begin{tabular}[c]{@{}c@{}}A5\\kin\end{tabular} & \begin{tabular}[c]{@{}c@{}}A6\\mar\end{tabular} & \begin{tabular}[c]{@{}c@{}}A7\\ary\end{tabular} & \begin{tabular}[c]{@{}c@{}}A8\\esp\end{tabular} & \begin{tabular}[c]{@{}c@{}}A9\\tel\end{tabular} & avg \\ 
    \hline
    baseline &  \textbf{.8300} & \textbf{\uline{.6900}} & \textbf{\uline{.7200}} & \textbf{\uline{.8800}} & \textbf{\uline{.7700}} & .7000 & .8200 & \textbf{\uline{.7729}} \\
    distilroberta-v1 (TranSem) & .8125 & .6403 & .6807 & .8406 & .7448 & \textbf{\uline{.7211}} & .8255 & .7522 \\
    \hline
     mpnet-base-v2   & .8104 & .6692 & .6971 & \textbf{.8568} & .7297 & .6518 & .8250 & .7486 \\
      roberta-large-v1   & \textbf{.8260} & .6750 & \textbf{.7056} & .8461 & .7480 & .6298 & \textbf{\uline{.8394}} & \textbf{.7528} \\
      sentence-t5-xl   & .8236 & .6440 & .6720 & .8324 & .7124 & .6277 & .8250 & .7339 \\
      multi-qa-mpnet-base-dot-v1  & .8111 & \textbf{.6852} & .7041 & .8482 & .7163 & .6586 & .8245 & .7497 \\
     all-MiniLM-L12-v2  & .8237 & .6474 & .7026 & .8522 & \textbf{.7605} & .6141 & .8247 & .7465 \\
     \hline
    FineSem-Individual  &.8385 & .6335 & \textbf{.7175} & .2211 & \textbf{.7647} & .6900 & .6085 & .6391 \\  %&.1551&.3434& 
    FineSem-Unified   & \textbf{\uline{.8438}} & .6369 & .6837 & .3878 & .6265 & \textbf{.7040} & .6993  &.6546\\  %&.2218&.3742& 
    FineSem-Translated & .8105 & \textbf{.6383} & .7133 & \textbf{.8608} & .7403 & .6663 & \textbf{.8152} & \textbf{.7493}  \\  %&.1731&.3392&     
    \hline
    \end{tabular}
}
}\caption{Model Performance (Spearman Correlation Coefficient) on Subtask A test set. $TranSem$ shows results submitted before the official deadline, $baseline$ shows official baseline results, and the rest are contrastive results for our various models.   The best scores within each section are \textbf{bolded}, and best scores across all sections are \textbf{\uline{underlined}}.}
\label{table:result_a}
\end{table*}

\subsubsection{Usefulness of Machine Translation and Direct Fine-tuning} \label{sec:a4}
We compare the performance of the \textit{FineSem} models fine-tuned using the 3 data options (results are shown in \cref{table:result_a}). FineSem-Individual shows the performance of T5-XL models fine-tuned on the individual datasets.  Unified and Translated models show the performance of the two T5-XL models fine-tuned on the untranslated data and translated$+$augmented data. The model trained on untranslated data performs poorly on the Marathi dataset, but performs on par with the other models indicating that we may not need to translate all languages to English. We find that direct fine-tuning with the translated and augmented data is comparable with the \textit{TranSem} model using various sentence embeddings.

\subsection{Track C Languages} \label{res:c}
In \cref{tab:finesem-c}, we compare the performance of various T5 models on the track C languages. We submitted official results using our T5-XL based FineSem model (FineSem-LB) where the results are obtained using the checkpoint after the third epoch.  With the same model we also report with the approach where we use the checkpoint which results in the best performance on the development data for a given language.  These results are shown as FineSem-XL.  We compare these results with the T5-base and T5-large based FineSem models.  We bold the best scores for easy readability but underline scores that are better than the baseline.  Among our models the overall performance of the XL model is better and this model improves upon the baseline for Afrikaans, Indonesian and Spanish.   

\begin{table*}[]
    \centering
    \resizebox{.98\textwidth}{!}{
    \begin{tabular}{|l|c|c|c|c|c|c|c|c|c|c|c|c|c|}
    \hline
   &C1&C2&C3&C4&C5&C6&C7&C8&C9&C10&C11&C12&\\     
   Model&Afr&Arq&Amh&Eng&Hau&Hin&Ind&Kin&Arb&Ary&Pan&Esp&avg\\
   \hline
   Baseline&.7900&\textbf{\uline{.4600}}&.\textbf{\uline{8400}}&\textbf{\uline{.8000}}&\textbf{\uline{.6200}}&\textbf{\uline{.7600}}&.4700&\textbf{\uline{.5700}}&\textbf{\uline{.6100}}&\textbf{\uline{.4000}}&\textbf{\uline{-.050}}&.6200&.5742\\
   FineSem-LB&\textbf{\uline{.8223}}&1263&.0430&.7875&.4569&.1552&\uline{.5153}&.4836&.0354&-.0375&-.0775&.6089&\textbf{.3266}\\
   \hline
   FineSem-XL&\uline{.8164}&1023&.0373&\textbf{.7889}&.4561&.1594&\textbf{\uline{.5279}}&.4128&.0000&.0219&-.0817&\uline{.6259}&.3246\\
   FineSem-L&.8007&-.0515&.0112&.7752&\textbf{.4831}&.1764&.4419&\textbf{.5094}&.0154&\textbf{.0331}&-.0591&.6605&.3164\\
   FineSem-B&.7802&\textbf{.1799}&\textbf{.2543}&.7448&.4784&\textbf{.2404}&.4517&.3861&\textbf{.0527}&.0268&\textbf{-.0520}&\textbf{\uline{.6289}}&.3477\\
   \hline
   \end{tabular}
   }
    \caption{Model performance (Spearman Correlation Coefficient) on Track C test sets. All language test sets (except English) use the FineSem models trained on the English training set. The English test set uses the FineSem models trained on the Spanish training set. The best scores among our models are \textbf{bolded}.  Scores better than baseline are \textbf{\uline{Underlined}}.}
\label{tab:finesem-c}

\end{table*}

\section{Related Work}
In this section, we present previous research conducted in the fields of Machine Translation (see \cref{rel:translation}), Sentence Embedding (see \cref{rel:sen_emb}) and Semantic Similarity (see \cref{rel:sts}).

\subsection{Machine Translation} \label{rel:translation}
%Early work in machine translation dates back to 1940s. However, with its recent migration from  statistical system to a transformer-based systemhas pushed it to a new boundary. \citet{bahdanau2014neural} first showed that an encoder-decoder-based seq-to-seq model can be developed for machine translation. Though a lot of research was done on high-resource languages, the number of systems developed for low-resource languages was very low until a few years back. The main challenge behind that was the performance of a neural network highly depends on the number of data \citep{haddow2022survey}. To counterattack this challenge, cross-lingual models were developed \citep{nguyen2017transfer, zoph2016transfer}. Some of the research has been done to develop datasets for low-resource languages \citep{banon2020paracrawl, schwenk2019ccmatrix}. Later \citet{costa2022no} has proposed several many-to-many models that outperform the state-of-the-art translation models by $\approx$ 40\%.

Machine translation has evolved in the last 75 years from rule-based systems to statistical-based systems to the current neural machine translation (NMT) systems. In the 10 years since the first sequence-to-sequence NMT model \citep{bahdanau2014neural}, machine translation reached a point where translations from models for high-resource languages rival human translators \citep{laubli-etal-2018-machine,popel}. This was possible due to the amount of bilingual data pairs available for training in these languages \citep{haddow2022survey}. Translation systems for medium and low resource languages that lacked the scale of these resources either developed cross-lingual models \citep{nguyen2017transfer, zoph2016transfer} or developed datasets \citep{banon2020paracrawl, schwenk2019ccmatrix}. Current state-of-the-art translation models use a many-to-many approach to handle a large number of medium to low-resource languages \citep{costa2022no}.

\subsection{Sentence Embedding} \label{rel:sen_emb}
%Generating sentence embedding can help from semantic searches to clustering. Keeping that in mind, a lot of effort was made to improve the sentence embedding. With the discovery of transformers \citep{vaswani2017attention}, encoder-only models have shown high performance in generating sentence embeddings, i.e., BERT \citep{devlin2018bert}, RoBERTa \citep{liu2019roberta}, XLNet \citep{yang-etal-2018-learning}. \citet{cer2018universal} has shown how learned sentence embeddings can be used in other NLP tasks. Later, \citet{Reimers_2019} provided multiple models that outperformed all the state-of-the-art models. It also has shown a way to extract similar sentences in a very short time (which was a bottleneck of BERT for semantic search systems). In their recent work \citep{reimers-2020-multilingual-sentence-bert}, authors have provided a teacher-student model that can efficiently be used to develop a system for a low-resource language. 

Generating a sentence-level embedding is useful for semantic searches and clustering. Since the first transformer model \citep{vaswani2017attention}, several encoder-only models such as BERT \citep{devlin2018bert}, RoBERTa \citep{liu2019roberta}, XLNet \citep{yang-etal-2018-learning} were used to learn effective sentence embeddings that also performed well on downstream NLP tasks \citep{cer2018universal, roy-dipta-etal-2023-semantically}. \citet{Reimers_2019} developed a Siamese-like network architecture with two BERT sentence embedding models that improved semantic search systems. Using T5 \citep{t5_raffel} models in a similar architecture, sentence embeddings produced by T5 were shown to be superior to the encoder-only model embeddings with performance gains in downstream tasks \cite{T5_ni}. A more recent work \citep{reimers-2020-multilingual-sentence-bert} showed that a teacher-student model can be efficiently used to develop a sentence embedding system for many low-resource languages. 

\subsection{Semantic Similarity} \label{rel:sts}
Semantic textual similarity captures a type of semantic relatedness requiring similarity on all aspects between a sentence pair. SemEval tasks on semantic textual similarity from 2012 to 2017 resulted in the STS benchmark \citep{cer-etal-2017-semeval}. Recently, \citet{deshpande2023csts} proposed conditional semantic textual similarity to explore semantic relatedness.
 
\section{Conclusion \& Future Work}

We developed two different models and showed how the models performed in supervised and cross-domain training tasks in 14 languages. We explored using machine translation, sentence encoders, and SST-B style training with T5 models. Our models improved over the official baseline for some of the languages. For computational purposes, we have excluded using more recent models like mistral-7b \citep{jiang2023mistral}, which have outperformed most of the open-source and close-source models in various benchmarks \citep{zheng2024judging}. For future work, we intend to explore prompting for STR and prompt-based LLMs \footnote{\url{https://chat.openai.com/}} for translation. 

\section{Disclaimer}
We did not use AI assistants to write any part of our paper or code.  All writing is original and produced by the authors.

\section{Limitations}
We acknowledge our work has the following limitations. We use several pre-trained LLMs in our experiments. It is well known that these models can echo biases and misinformation either implicitly or explicitly. We did not control for any of these when training them on the STR datasets. In addition, the STR datasets may also echo several biases related to social groups, cultural groups, race, gender, behavioral, and perceptual differences of annotators. We did not explore or control for any of these biases in our work. As a result, our work carries the limitations of both the models and the datasets we used.

\bibliography{custom}

% \appendix

% \section{Example Appendix}
% \label{sec:appendix}

% This is an appendix.

\end{document}